\documentclass{bmvc2k}

\def\thefootnote{$\dagger$}\footnotetext{This work is done at Bosch Center for Artificial Intelligence.}

\usepackage{times}
\usepackage{amsmath}
\usepackage{amsthm}
\usepackage{amssymb}
\usepackage{makecell}
\usepackage{array}
\usepackage{multirow}
\usepackage{graphicx}
\usepackage[font=small]{caption}
\usepackage{wrapfig}

\def\etal{\emph{et al}.}

\title{TaylorSwiftNet: Taylor Driven Temporal Modeling for Swift Future Frame Prediction}

\addauthor{Saber Pourheydari*}{m.saberpourheydari@gmail.com}{1}
\addauthor{Emad Bahrami*}{bahrami@iai.uni-bonn.de}{1}
\addauthor{Mohsen Fayyaz*}{mohsenfayyaz@microsoft.com}{1, 2}
\addauthor{Gianpiero Francesca}{gianpiero.francesca@toyota-europe.com}{3}
\addauthor{Mehdi Noroozi\thefootnote{}}{m.noroozi@samsung.com}{4}
\addauthor{Juergen Gall}{gall@iai.uni-bonn.de}{1}
\addinstitution{
 Computer Vision Group\\
 University of Bonn\\
 Bonn, Germany
}
\addinstitution{
 Microsoft\\
 Berlin, Germany
}
\addinstitution{
Toyota Motor Europe\\
 Brussels, Belgium\\
}
\addinstitution{
Samsung AI\\
 Cambridge, UK\\
 \vspace{2mm}
 * indicates equal contribution
}

\runninghead{Pourheydari \etal}{TaylorSwiftNet}

\begin{document}
\maketitle

\begin{abstract}
While recurrent neural networks (RNNs) demonstrate outstanding capabilities for future video frame prediction, they model dynamics in a discrete time space, i.e., they predict the frames sequentially with a fixed temporal step. 
RNNs are therefore prone to accumulate the error as the number of future frames increases. In contrast, partial differential equations (PDEs) model physical phenomena like dynamics in a continuous time space. However, the estimated PDE for frame forecasting needs to be numerically solved, which is done by discretization of the PDE and diminishes most of the advantages compared to discrete models.   
In this work, we, therefore, propose to approximate the motion in a video by a continuous function using the Taylor series. 
To this end, we introduce TaylorSwiftNet, a novel convolutional neural network that learns to estimate the higher order terms of the Taylor series for a given input video. TaylorSwiftNet can swiftly predict future frames in parallel and it allows to change the temporal resolution of the forecast frames on-the-fly. The experimental results on various datasets demonstrate the superiority of our model.
\end{abstract}

\section{Introduction}
\label{introduction}

The ability to predict future frames of a video is essential for many applications such as weather forecasting \cite{convolutionalLSTM}, autonomous driving \cite{kwon2019predicting}, robotics \cite{finn2016unsupervised}, or action recognition \cite{liang2017dual}. When only the raw video is given, the task is very challenging since it requires to learn the complex motion of the objects present in the video. To address this task, several approaches have been proposed over the last years. In particular,  auto-regressive methods and recurrent neural networks (RNNs) have been popular~\cite{convolutionalLSTM, predrnn, predrnn++, 3DLSTM}. 

While these approaches learn the motion implicitly, recently a new line of work appeared that leverages partial differential equations (PDEs) and deep learning for forecasting video frames \cite{PDE-Net, PDE-Net2, PhyDNet, Data-Driven-PDEs, raissi2018deep}. PDEs are very appealing for this task since they are an appropriate tool to model physical phenomena like dynamics. These methods model motion in the continuous time space in contrast to auto-regressive models or  recurrent neural networks that model motion sequentially in a discrete time space. 
This has two major advantages. First, the motion model is independent of the sampling rate and it is, for instance, possible to forecast the motion at a higher sampling rate than the observed frames. Second, which is more important, a continuous PDE can be solved for any future point. 
It is therefore not required to sequentially go through all frames until the desired point in the future is reached, which increases the inference time for points that are more distant in the future and which is prone to error accumulation.
\newpage
The estimated PDEs, however, do not provide directly the prediction, but they need to be numerically solved, which diminishes most of the advantages compared to discrete models.

\begin{figure}[t]
\begin{center}
    \includegraphics[trim={0 10pt 0 0 }, clip, width=0.85\columnwidth]{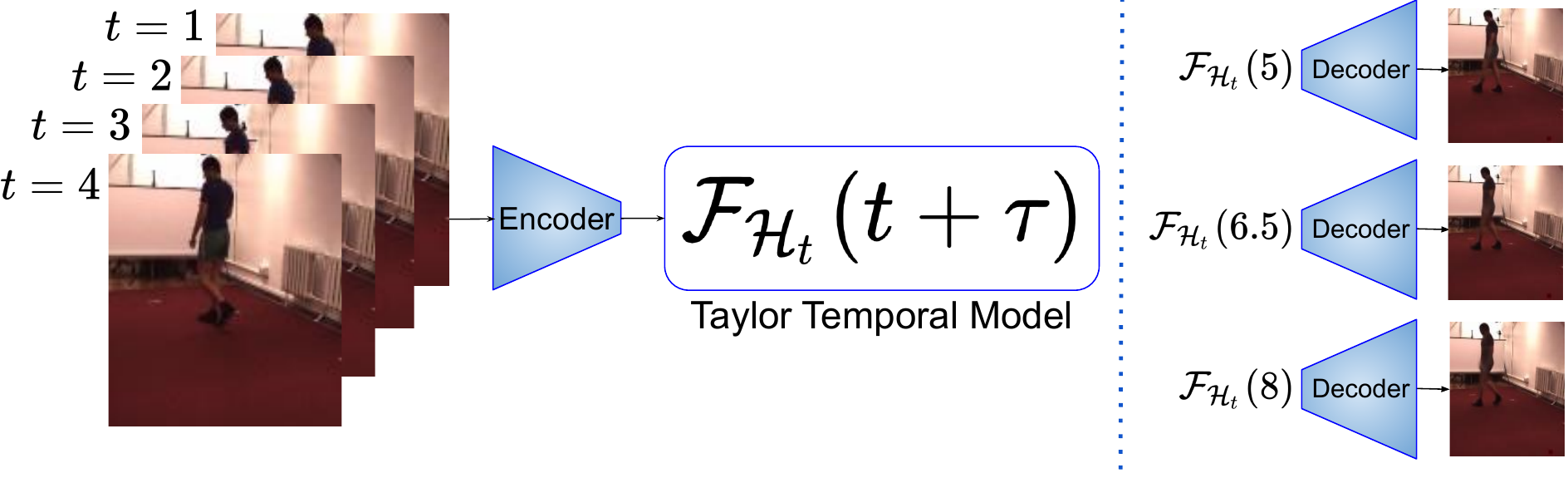}    
\end{center}
\caption{Given a sequence of observed frames until time $t$, the encoder maps them into a latent space. Our network infers from the observations in the latent space $\mathcal{H}_t$, a continuous function $\mathcal{F}_{\mathcal{H}_t}(t+\tau)$ in one forward pass. The inferred function can then be evaluated for any positive value $\tau$ in order to forecast future frames at $t+\tau$. This can be done in parallel for different values of $\tau$ and the decoder maps the forecast frames back to the image domain. 
}
\label{fig:teaser}
\end{figure}

In this work, we, therefore, propose a novel approach that takes full advantage of a continuous representation of motion. Instead of learning a PDE that needs to be numerically solved for training and inference, we directly infer a continuous function that describes the future. In contrast to RNNs that forecast the future frame-by-frame or PDE-based approaches that discretize PDEs to solve them numerically, we infer a continuous function over time from the observations. This avoids discretization artifacts and provides an analytical function that can be swiftly evaluated for any future continuous point as illustrated in Fig.~\ref{fig:teaser}. 
This allows, for instance, to generate frames at different future points in parallel. We can also forecast future frames at a higher sampling rate than the observed frames. All these properties are very useful for practical applications and demonstrate the advantages of a continuous representation.                                      

It is, however, very challenging to infer such a continuous function from a discrete set of high-dimensional observations as it is the case for video frame forecasting. We, therefore, propose to approximate the unknown function by the Taylor series around the last observation $t$ up to a finite order. As illustrated in Fig.~\ref{fig:teaser}, we first map the observed frames into a learned embedding space $\mathcal{H}_t\in\mathcal{H}$ and then estimate each term of the Taylor expansion $\mathcal{F}_{\mathcal{H}_t}$. An important aspect of the proposed TaylorSwiftNet is that it learns to generate a full Taylor expansion from a discrete set of observations, i.e., after training, the network is capable of inferring a Taylor expansion of an unknown function only from a set of observations until point $t$. We validate the capabilities of the network using simulated data of known functions where the ground-truth Taylor expansion is known as well for high-dimensional problems like video frame forecasting as shown in Fig.~\ref{fig:teaser}. Since the network infers a full function over time $\mathcal{F}_{\mathcal{H}_t}: \mathbb{R} \mapsto \mathcal{H}$, we can evaluate the function for any real value $t{+}\tau\in\mathbb{R}$. For instance, we can swiftly generate frames for $\tau =1, 1.5, 4$ as in Fig.~\ref{fig:teaser}. Since the predictions are in the embedding space, the predictions are mapped back to the image domain by the decoder. The entire TaylorSwiftNet consisting of the encoder, the estimation of the Taylor expansion, and the decoder is trained end-to-end. 

We compare the proposed approach with state-of-the-art methods on four datasets from different application domains ranging from video frame forecasting to forecasting sea surface temperature. On all datasets, TaylorSwiftNet outperforms RNN-based as well as PDE-based approaches. We furthermore demonstrate the capabilities of the approach, namely estimating Taylor expansions of analytical functions and the flexibility to forecast frames with a higher frame rate than the observed data.

\section{Related Work}
\label{related_work}
Several approaches have been proposed for video forecasting using unlabeled videos. In particular, deep neural networks have shown promising results for this task. \cite{patraucean2015spatio, luo2017unsupervised, li2018flow} use optical flow to model changes of temporal dynamics and to better predict the future. Some works \cite{2016visual, dynamicfilternet, finn2016unsupervised} focus on modeling the geometric transformations between frames to predict future frames. \cite{farazi2019frequency} propose to solve the video prediction task by estimating and using the transformations of the signal in the frequency domain.
\cite{soft-dtw, vincent2019shape} focus on improving the sharpness of the predicted frames by using custom loss functions. To better handle the future uncertainty and generate sharp predictions, \cite{beyond_mse, vondrick2016generating, kwon2019predicting, SVG} have used generative adversarial networks or variational autoencoders.
Methods based on 2D or 3D CNNs \cite{contextvp, beyond_mse, vondrick2016generating, Gao2022} have also been proposed. In particular recurrent neural networks (RNNs) have been popular \cite{unsupLSTMs, predrnn, predrnn++, finn2016unsupervised, Lu_2017_CVPR, Oliu_2018_ECCV, memory19, convolutionalLSTM, 3DLSTM} in recent years.

While these approaches learn the motion implicitly, recently a new line of work appeared that leverages partial differential equations (PDEs) and neural networks for forecasting video frames \cite{dona2020pde, PDE-Net, PDE-Net2, PhyDNet, Data-Driven-PDEs, raissi2018deep}.
Some recent PDE-based works demonstrate substantial improvements compared to recurrent neural networks for video frame forecasting.
\cite{PhyDNet, rubanova2019latent, ayed2020learning, yildiz2019ode2vae} define the dynamics using learned ordinary differential equations following \cite{chen2018neural}.
\cite{ryder2018black, li2020scalable, franceschi2020stochastic} employ differential equations for stochastic data.
Some methods shape the prediction function or the cost function of their methods using prior physical knowledge \cite{brunton2016discovering, de2018end}. For instance, \cite{de2019deep} uses general advection-diffusion principles as a guideline for designing a network. \cite{brunton2016discovering, rudy2017data, schaeffer2017learning} discover the PDEs by sparse regression of potential differential terms. 
\cite{hamiltonian, chen2019symplectic, toth2019hamiltonian} introduce non-regression loss functions inspired by Hamiltonian mechanics \cite{hamilton1835vii}.
\cite{qin2019data, fablet2018bilinear} have designed specific architectures for predicting and identifying dynamical systems inspired by numerical schemes for solving PDEs and residual neural networks \cite{chen2018neural, lu2018beyond, li2019deep, zhu2018convolutional}. \cite{dona2020pde} proposes the separation of variables as a general paradigm based on a resolution method for partial differential equations for video prediction and disentanglement.
PDE-Net \cite{PDE-Net,PDE-Net2} discretizes a broad class of PDEs by approximating partial derivatives with convolutions.

Although partial differential equations model the motion in the continuous time space, the PDE-based approaches discretize the PDEs using, for instance, the forward Euler method. In this work, we propose a different continuous representation that does not need any	discretization.

\section{Forecasting Future Frames}
The problem of forecasting future frames can be formulated as 
\begin{equation}\label{eq:pred}
    p(x_{t + 1} | \mathcal{X}_t) \quad \text{for}\quad \mathcal{X}_t = \{x_{t-k},\dots,x_{t}\},
\end{equation}
where the probability of a future frame $x_{t+1}$ is conditioned on the past $k$ observed frames.
For forecasting with a different temporal step, i.e., $\tau > 1, \, \tau \in \mathbb{N}$, this results in   
\begin{align}
    p(x_{t + \tau} | \mathcal{X}_t) 
     & = \int_{x_{t + \tau - 1}} \dots \int_{x_{t + 1}} p(x_{t + \tau} | x_{t + \tau - 1}, \mathcal{X}_t)  \dots  \nonumber\\
    &  p(x_{t + 2} | x_{t + 1}, \mathcal{X}_t) p(x_{t + 1} | \mathcal{X}_t)
    dx_{t + \tau - 1} \dots dx_{t + 1}. \nonumber
\end{align}
When the motion is complex and $x_{t+\tau}$ is high-dimensional, as it is the case for video frame forecasting, the computation of the integrals is infeasible. Common auto-regressive approaches therefore approximate the solution frame-by-frame by taking the argmax of \eqref{eq:pred} and adding the new estimate $x_{t+1}$ to the observations. This, however, has the disadvantage that one needs to iterate over all frames until $x_{t+\tau}$ is reached, the approximation error increases over time, and $\tau$ is constrained by the frame-rate of the training and observed data.   

To overcome these issues, we propose to learn a mapping from the space of observations $\mathcal{X}$ to the space of infinitely differentiable functions $\mathcal{F}: \mathbb{R} \mapsto \mathcal{X}$. Note that we learn the mapping to a function space and not the Euclidean space $\mathcal{X}$. During inference, the learned mapping maps a given observation $\mathcal{X}_t$ to a continuous forecasting model $\mathcal{F}_{\mathcal{X}_t}$, which can then be evaluated for any $\tau\in\mathbb
{R}_{>0}$:   
\begin{equation}
    x_{t + \tau} = \mathcal{F}_{\mathcal{X}_t}(t + \tau)
    \label{eq:euler}
\end{equation}
as it is illustrated in Fig.~\ref{fig:teaser}. This means that we can forecast a frame in a more distant future directly without forecasting all intermediate frames. It allows forecasting all future frames in parallel instead of forecasting them sequentially. Finally, we can even forecast at super temporal resolution, i.e., at a higher frame-rate than the observations, without the need to re-train the model.

\section{Temporal Dynamics Modeling Through Taylor Series}\label{sec:Taylor}
\begin{figure*}[t]
\begin{center}
    \includegraphics[trim={1.2cm 0 1.0cm 0.3cm}, clip, width=1.0\linewidth]{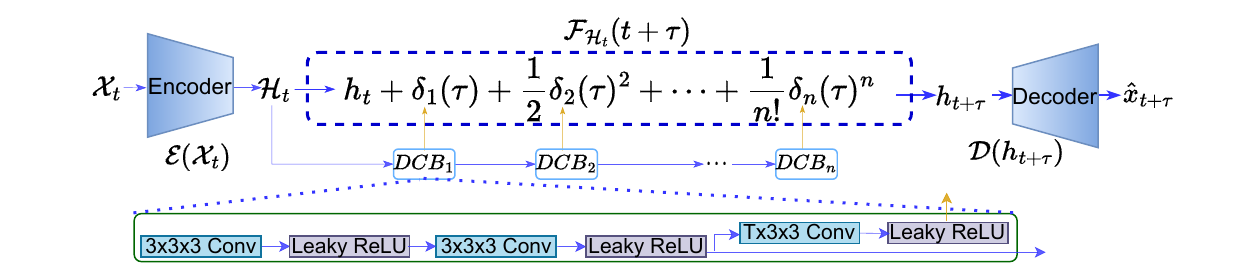}
\end{center}
\caption{TaylorSwiftNet. The network gets a sequence of frames $\mathcal{X}_t \in \mathbb{R}^{C \times T \times H \times W }$ as input. The encoder maps all frames into a latent space. The frames in the latent space are denoted by $\mathcal{H}_t \in \mathbb{R}^{C'\times T\times H'\times W'}$ and $h_t \in \mathbb{R}^{C'\times H'\times W'}$ denotes the embedding for frame $t$. Using a Taylor series of order $n$, the function $\mathcal{F}_{\mathcal{H}_t}(t+\tau)$ is approximated to model the temporal dynamics. To estimate the higher order terms of the Taylor series,
$\mathcal{H}_t$ is fed to the DC blocks. The DC blocks estimate sequentially the $\delta_i$s, which are the estimated derivatives of $\mathcal{F}_{\mathcal{H}_t}$ at the point $t$. Having the derivatives $\delta_i$, the Taylor series approximates $\mathcal{F}_{\mathcal{H}_t}(t+\tau)$ at the future time step $\tau$, which yields $h_{t+\tau}$. Finally, $h_{t+\tau}$ is fed to the decoder and the future frame $\hat{x}_{t+\tau}$ is predicted. 
}
\label{fig:TaylorSwiftNet}
\end{figure*}

As discussed, our novel approach for continuous forecasting can directly predict $\hat{x}_{t+\tau}$ for any value $\tau\in\mathbb
{R}_{>0}$ from the observations $\mathcal{X}_t$. Since using the observed frames $\mathcal{X}_t$ directly is not practical, we embed the observed frames in an embedding space $\mathcal{H}_t$ and infer the continuous motion model \eqref{eq:euler} in the embedding space, i.e.,  $\mathcal{F}_{\mathcal{H}_t}: \mathbb{R} \mapsto \mathcal{H}$ where $\mathcal{F}_{\mathcal{H}_t}$ is infinitely differentiable and smooth. As illustrated in Fig.~\ref{fig:TaylorSwiftNet}, the entire model thus consists of three parts that are learned end-to-end:
\begin{equation}
    \hat{x}_{t+\tau} = \mathcal{D}(h_{t+\tau})\:;\:
    h_{t+\tau} = \mathcal{F}_{\mathcal{H}_t}(t+\tau)  \:;\:
    \mathcal{H}_t = \mathcal{E}(\mathcal{X}_t).
    \label{eq:general_model}
\end{equation}

$\mathcal{E}(.)$ first maps the observed frames $\mathcal{X}_t$ into the learned embedding space. Given the embedding of video frames represented by $\mathcal{H}_t$, our network infers an observation specific function $\mathcal{F}_{\mathcal{H}_t}(.)$ that models the future dynamics in the embedding space. It is important to note that our approach infers a full function with respect to $\tau$ and not a single point estimate as it is done by standard networks. The function $\mathcal{F}_{\mathcal{H}_t}(.)$ can therefore be swiftly evaluated for any value. $\mathcal{D}(.)$ finally decodes the forecast embedding $h_{t+\tau}$ and predicts the future frame $\hat{x}_{t+\tau}$. While $\mathcal{E}(.)$ and $\mathcal{D}(.)$ will be discussed in Section \ref{sec:arch}, we first discuss $\mathcal{F}_{\mathcal{H}_t}$.

Our goal is to learn the unknown continuous function $\mathcal{F}_{\mathcal{H}_t}$ that represents the future dynamics in the embedding space. In case of $\tau=0$, $\mathcal{F}_{\mathcal{H}_t}(t) = h_t$, which is equal to the last vector of $\mathcal{H}_t$. For $\tau>0$, however, this function is very complex. We, therefore, propose to approximate the function using the Taylor series:
\begin{equation}
    \mathcal{F}_{\mathcal{H}_t}(t+\tau) \simeq \sum_{n=0}^{\gamma} \frac{\mathcal{F}_{\mathcal{H}_t}^{(n)}(t)}{n!}\tau^n \:;\: \mathcal{F}_{\mathcal{H}_t}^{(n)} = 
    \frac{\partial^n \mathcal{F}_{\mathcal{H}_t}}{\partial \tau^n}.
    \label{eq:taylor}
\end{equation}
If $\mathcal{F}_{\mathcal{H}_t}(t+\tau)$ is an analytic function, the Taylor series with $\gamma=\infty$ is equal to $\mathcal{F}_{\mathcal{H}_t}(t+\tau)$. In practice, we approximate it by using only a finite number of terms. 

Although the approximation \eqref{eq:taylor} requires to compute $\mathcal{F}_{\mathcal{H}_t}^{(n)}$ for higher order terms at $t$ in order to get a good approximation, this needs to be done only once and the function can be evaluated for any $\tau>0$. Since computing the derivatives of an unknown function in a very high dimensional space is impractical, we propose to learn a network that infers them from $\mathcal{H}_t$:

\begin{equation}
    \mathcal{F}_{\mathcal{H}_t}^{(n)}(t) \simeq f_n\left(\Delta_n\left(\mathcal{H}^{(n-1)}_t\right)\right);\; \mathcal{H}^{(n-1)}_t = \Delta_{n-1}\left(\mathcal{H}^{(n-2)}_t\right) 
    \label{eq:DerivativeEstimators}
\end{equation}
where $f_n$ and $\Delta_n$ are trainable blocks that will be described in Section~\ref{sec:arch} and $\mathcal{H}^{(0)}_t = \mathcal{H}_t$. Considering \eqref{eq:taylor} and \eqref{eq:DerivativeEstimators}, we can reformulate \eqref{eq:general_model} as:
\begin{equation}
    \hat{x}_{t+\tau} =  \mathcal{D}\left(h_t + \sum_{n=1}^{\gamma} \frac{f_n(\Delta_{n}(\Delta_{n-1}(\ldots \Delta_{1}( \mathcal{E}(\mathcal{X}_t)))))}{n!}\tau^n\right). 
    \label{eq:full}
\end{equation}
Note that the $\gamma$ terms of the Taylor series are computed recursively, but only once for a given observation. All future frames $t+\tau$ can be predicted in parallel as it is illustrated in Fig.~\ref{fig:teaser}.

Having the future frame $x_{t+\tau}$ as ground-truth during training, we can compute the loss between the predicted frame $\hat{x}_{t+\tau}$ and the ground-truth frame $x_{t+\tau}$ and update the parameters of $\mathcal{D}$, $f_n$, $\Delta_{n}$, and $\mathcal{E}$.


\section{Proposed Architecture}\label{sec:arch}
We now describe the network architecture that learns the model formulated in \eqref{eq:full}. As illustrated in Fig.~\ref{fig:TaylorSwiftNet}, the network first encodes the input video frames, i.e., $\mathcal{H}_t = \mathcal{E}(\mathcal{X}_t)$. 
Then the temporal model approximates the function $\mathcal{F}_{\mathcal{H}_t}(t+\tau)$, and finally, we only need to apply the decoder $\mathcal{D}$ to the output of the temporal model $\mathcal{F}_{\mathcal{H}_t}(t+\tau)$ to generate images for all relevant positive values $\tau$ in parallel, as illustrated in Fig.~\ref{fig:teaser}.

\textbf{Encoder.}
The encoder $\mathcal{E}(\mathcal{X}_t)$ maps the input video frames $\mathcal{X}_t \in \mathbb{R}^{C \times T \times H \times W }$ to $\mathcal{H}_t \in \mathbb{R}^{C'\times T\times H'\times W'}$. $C, T, H, W$ are the video channels, number of frames, height, and width of the frames. $C', H', W'$ are the channels of the feature maps and their height and width. We have used a modified version of 3DResNet \cite{hara2017learning, he2016deep} for the encoder, which is described in the Appendix, but any 3D convolutional neural network can be used in principle. 

\textbf{Temporal Model.}
The temporal model $\mathcal{F}_{\mathcal{H}_t}(t+\tau)$ models the 
future temporal dynamics and forecasts the frame at the future temporal step $t{+}\tau \in \mathbb{R}$ in 
the embedded space $h_{t+\tau}\in \mathbb{R}^{C'\times H' \times W'}$. As illustrated in Fig.~\ref{fig:TaylorSwiftNet}, we estimate $\mathcal{F}_{\mathcal{H}_t}^{(n)}$ \eqref{eq:DerivativeEstimators} recursively. We use for each $\Delta_n(.)$ a convolutional block called delta convolutional block (DCB). The first 2 convolutional layers of DCB use kernels with size $3\times3\times3$ and stride $1\times1\times1$. The input and output size remains the same. 

The output feature map is then fed to: (a) the final convolutional layer $f_n$ to output the estimated derivative $\delta_n\in \mathbb{R}^{C'\times H'\times W'}$; (b) the next DC block to estimate the next order derivative $\delta_{n+1}$. $f_n$ uses kernels with size $T\times3\times3$. The estimated derivatives are then used in \eqref{eq:DerivativeEstimators} to model the temporal dynamics and forecast the embedding $h_{t+\tau} \in \mathbb{R}^{C'\times H'\times W'}$. While in our experiments the different DC blocks do not share their weights, we also evaluate a recurrent version with shared weights in the Appendix. 

\textbf{Decoder.}
The decoder is a convolutional neural network that consists of 6 convolutional layers with kernel size $1\times3\times3$. 
The decoder $\mathcal{D}(h_{t+\tau})$ decodes the embedding and predicts the future frame $\hat{x}_{t+\tau} \in \mathbb{R}^{C\times H\times W}$. For more details regarding the implementation details of the decoder, we refer to the Appendix.


\section{Experiments}

\subsection{Datasets and Evaluation Metrics}
Following the state-of-the-art \cite{PhyDNet}, we also evaluate our method on four datasets from very different domains, namely Moving MNIST \cite{unsupLSTMs}, Human 3.6M \cite{human36}, Traffic BJ \cite{trafficBJ}, and Sea Surface Temperature \cite{seasurface}.

\textbf{Moving MNIST} is a standard dataset for sequence prediction which consists of two random digits moving inside a $64 \times 64$ grid. Data for training were generated on the fly and a test set of 10,000 sequences was used for evaluation. We predict 10 unseen future frames given 10 seen input frames.

\textbf{Human 3.6M} contains human actions with their corresponding 3D poses for 17 action scenarios. Following the setting of \cite{PhyDNet, memory19}, we select subjects S1, S5, S6, S7, and S8 for training and subjects S9 and S11 for testing using the walking action. Human 3.6M includes originally RGB images of size $1000 \times 1000 \times 3$ which we resize to $ 128 \times 128 \times 3$ for our experiments. We predict 4 unseen frames given 4 input seen frames.       

\textbf{Traffic BJ} contains the hourly taxi flows of Beijing in a $32 \times 32$ grid. Each frame has two channels corresponding to the traffic flow entering and leaving a district. We use 4 input seen frames to predict 4 unseen frames.

\textbf{Sea Surface Temperature} consists of meteorological data of the Atlantic ocean generated by NEMO (Nucleus for European Modeling of the Ocean), which is a state-of-the-art simulation engine for modeling ocean dynamics. 
Following the protocol of \cite{seasurface}, we use the Sea Surface Temperature (SST) data of $64 \times 64$ sized sub-regions extracted from the original $481 \times 781$ sized data. We predict 4 future frames given 4 unseen input frames.      

\textbf{Evaluation Metrics.} Following the state-of-the-art methods \cite{PhyDNet, PDE-Net, predrnn++}, we use the following evaluation metrics: Mean Squared Error (MSE), Mean Absolute Error (MAE), and the Structural Similarity (SSIM) \cite{structuralsimilarity}. We average the metrics over all frames of the predicted output sequence. While lower MSE and MAE indicate better performance, a higher SSIM is better.

If not otherwise specified, we use a Taylor model of order 4 for the Moving MNIST dataset and of order 2 for the other datasets. We refer to the Appendix for more implementation details. 

\begin{table*}[t]
\begin{center}
    \resizebox{1.0\columnwidth}{!}{
        \begin{tabular}{|c|ccc|ccc|ccc|ccc|}
            \Xhline{2\arrayrulewidth}
             & \multicolumn{3}{|c|}{\textbf{Moving MNIST}} & \multicolumn{3}{|c|}{\textbf{Traffic BJ}} & \multicolumn{3}{|c|}{\textbf{Sea Surface Temperature}} & \multicolumn{3}{|c|}{\textbf{Human 3.6M}}  \\
             \hline
             Method & MSE $\downarrow$ & MAE $\downarrow$ & SSIM $\uparrow$ & MSE $\times 100$ & MAE & SSIM & MSE $\times 10$ & MAE & SSIM & MSE $/ 10$ & MAE $/ 100$ & SSIM \\
             \hline 
             Advection-diffusion \cite{de2019deep} & - & - & - & - & - & - & 34.1 & 54.1 & 0.966 & - & - & - \\
             DDPAE \cite{ddpae} & 38.9 & 90.7 & 0.922 & - & - & - & - & - & - & - & - & - \\
             ConvLSTM \cite{convolutionalLSTM} & 103.3 & 182.9 & 0.707 & 48.5 & 17.7 & 0.978 & 45.6 & 63.1 & 0.949 & 50.4 & 18.9 & 0.776 \\
             PredRNN \cite{predrnn} & 56.8 & 126.1 & 0.867 & 46.4 & 17.1 & 0.971 & 41.9 & 62.1 & 0.955 & 48.4 & 18.9 & 0.781 \\
             Causal LSTM \cite{predrnn++} & 46.5 & 106.8 & 0.898 & 44.8 & 16.9 & 0.977 & 39.1 & 62.3 & 0.929 & 45.8 & 17.2 & 0.851 \\
             MIM \cite{memory19} & 44.2 & 101.1 & 0.910 & 42.9 & 16.6 & 0.971 & 42.1 & 60.8 & 0.955 & 42.9 & 17.8 & 0.790 \\ 
             E3D-LSTM \cite{3DLSTM} & 41.3 & 86.4 & 0.920 & 43.2 & 16.9 & 0.979 & 34.7 & 59.1 & 0.969 & 46.4 & 16.6 & 0.869 \\
             PhyDNet \cite{PhyDNet} & 24.4 & 70.3 & 0.947 & 41.9 & 16.2 & 0.982 & 31.9 & 53.3 & 0.972 & 36.9 & 16.2 & 0.901 \\
              SimVP \cite{Gao2022} & 23.8 & 68.9 & 0.948 & 41.4 & 16.2 & 0.982 & - & - & - & 31.6 & \textbf{15.1}  & 0.904 \\
             \hline
             TaylorSwiftNet (ours) & \textbf{17.8} & \textbf{42.5} & \textbf{0.965} & \textbf{35.3} & \textbf{13.7} & \textbf{0.992} & \textbf{29.8} & \textbf{52.2} & \textbf{0.978} & \textbf{23.1} & 15.8 & \textbf{0.910} \\
             \Xhline{2\arrayrulewidth}
        \end{tabular}
    }    
\end{center}
\caption{Comparison to the state-of-the-art on four datasets. The results are the mean over all predicted frames. 
}
    \label{tab:sta}
\end{table*}

\subsection{Comparison to State-of-the-Art}
\begin{figure*}[t]
\setlength{\textfloatsep}{5pt plus 1.0pt minus 2.0pt}
\begin{center}
\includegraphics[trim={25pt 70pt 15pt 0}, clip, width=0.9\columnwidth]{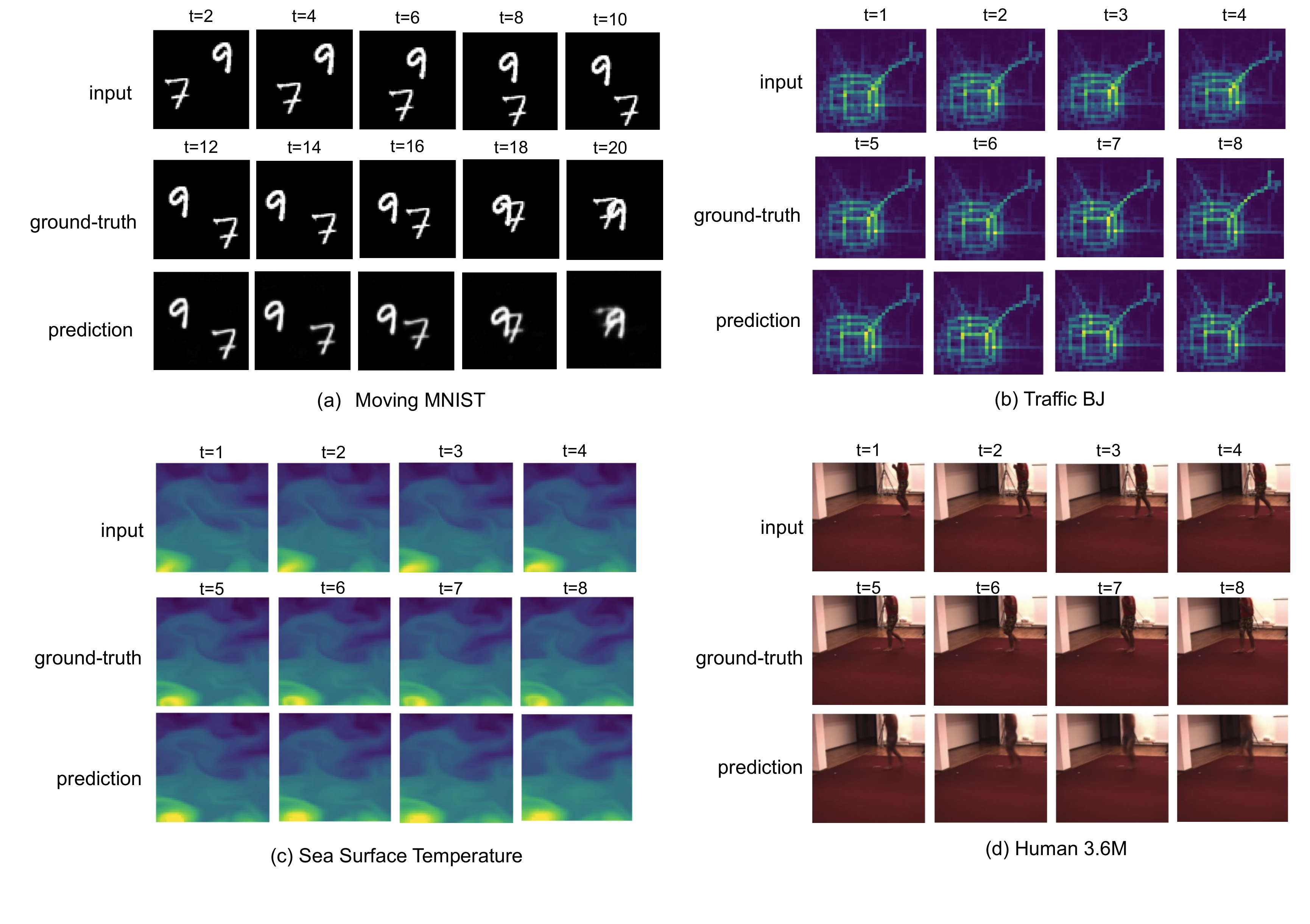}    
\end{center}
\caption{Qualitative results. For each dataset, the first row shows the input of the model, the second row shows the ground-truth, and the third row shows our prediction.
}
\label{fig:qualitatives}
\end{figure*}
We compare our TaylorSwiftNet with various state-of-the-art methods. As it can be seen in Table~\ref{tab:sta}, TaylorSwiftNet significantly outperforms the state-of-the-art methods on all  datasets and for all metrics. Only the very recent CNN-based approach SimVP~\cite{Gao2022} achieves a slightly lower MAE on Human3.6M. Our direct future temporal forecasting method significantly outperforms state-of-the-art architectures based on PDEs or recurrent neural networks such as PhyDNet~\cite{PhyDNet}, ConvLSTM~\cite{convolutionalLSTM}, PredRNN~\cite{predrnn}, Causal LSTM~\cite{predrnn++}, or Memory in Memory (MIM)~\cite{memory19}. 

For the Human 3.6M dataset, the approach \cite{villegas2017learning} uses additional supervision like human poses. Even without this additional supervision, our approach achieves a Peak Signal over Noise Ratio (PSNR) of 25.75 while \cite{villegas2017learning} reports in the appendix for time step 4 a PSNR below 21 and around 22 for the sequences with the least human motion. 
The results demonstrate that the proposed approach learns a very good temporal model from various videos of different application domains.

Fig.~\ref{fig:qualitatives} shows some qualitative results for all datasets. The qualitative results demonstrate the high quality of the forecast results that are generated by the proposed TaylorSwiftNet. For Moving MNIST, TaylorSwiftNet predicts accurately the digits even when they overlap. For Traffic BJ, the hourly taxi flows are correctly predicted. 
The sea surface temperature, which depends on phenomena that can be described by PDEs, is accurately predicted as well. And finally, TaylorSwiftNet also precisely anticipates the future position and pose of the person in the video from the Human 3.6M dataset.

\subsection{Ablation Experiments}
For the ablation studies, we compare the approach to different baselines, analyze the impact of the order of the Taylor series ($\gamma$), the performance for long-term forecasting, and the ability to forecast at a higher frame-rate than the observation without re-training the model. Furthermore, we analyze the accuracy of the learned terms of the Taylor expansion in the Appendix. 

\textbf{Comparison to Baselines.}
To demonstrate that the accuracy is due to the proposed model that infers a continuous function over $\tau$ from the observations and not due to the encoder and decoder, we compare our approach with two variants that use the same encoder and decoder. However the baseline models use $\tau$ as a conditioning variable, i.e., they use $\tau$ as input and generate a point estimate for a single $\tau$.

The \textit{Flatten} approach flattens the given hidden embedding $\mathcal{H}_t$ using three convolutional layers with kernel size of $3\times 3\times 3$, concatenates it to $v_{\tau}$ and afterwards up-samples it to obtain $h_{t+\tau}$ by using three transposed convolutional layers with kernel size of $1\times 3\times 3$. 
The \textit{Expand} approach expands the vector $v_{\tau}$ to the tensor $C'\times H'\times W'$ and concatenates it with the tensor from $\mathcal{H}_t$ which is temporally squeezed. By concatenating these two tensors channel-wise, we get a $2C'\times H'\times W'$ tensor. With one convolutional layer, we can down-sample the channels to estimate $h_{t+\tau}$. Both approaches are visualized in the Appendix. In addition, we evaluate TaylorSwiftNet using numerical derivatives instead of DC blocks. We numerically calculate the derivatives for the Taylor terms over the embeddings, i.e.,  $\delta_1=h_{t}-h_{t-1}$ and $\delta_2=\delta_1 - (h_{t-1} - h_{t-2})$ where $\delta_1$ and $\delta_2$ are the first and second order derivatives. Even in this setup, we train the model end-to-end. 

The results in Table \ref{tab:temporal_baseline} show that learning a function for all values of $\tau$ performs better than adding $\tau$ as input to the network and that the DC blocks perform better than the numerical derivatives.   

\begin{table}[t]
\begin{center}
        \begin{tabular}{|c|c|c|c|c|}
            \Xhline{2\arrayrulewidth}
             Method & MSE & MAE & SSIM  \\
            \hline 
            Point Estimate (Expand)  & 45.1 & 69.4 & 0.642  \\
            \hline
            Point Estimate (Flatten) & 43.6 & 63.6 & 0.887  \\
            \hline
            Numerical Derivatives & 23.1 & 46.3 & 0.956  \\
            \hline
            TaylorSwiftNet & \textbf{17.8} & \textbf{42.5} & \textbf{0.965}\\
            \Xhline{2\arrayrulewidth}
        \end{tabular}    
\end{center}
\caption{Comparison of the proposed temporal model to three variants on Moving MNIST.}
\label{tab:temporal_baseline}
\end{table}

\begin{figure}[t]
\centering
\subfloat[][]{
\centering
    \includegraphics[trim={0 0 0 6pt}, clip, width=0.38\columnwidth]{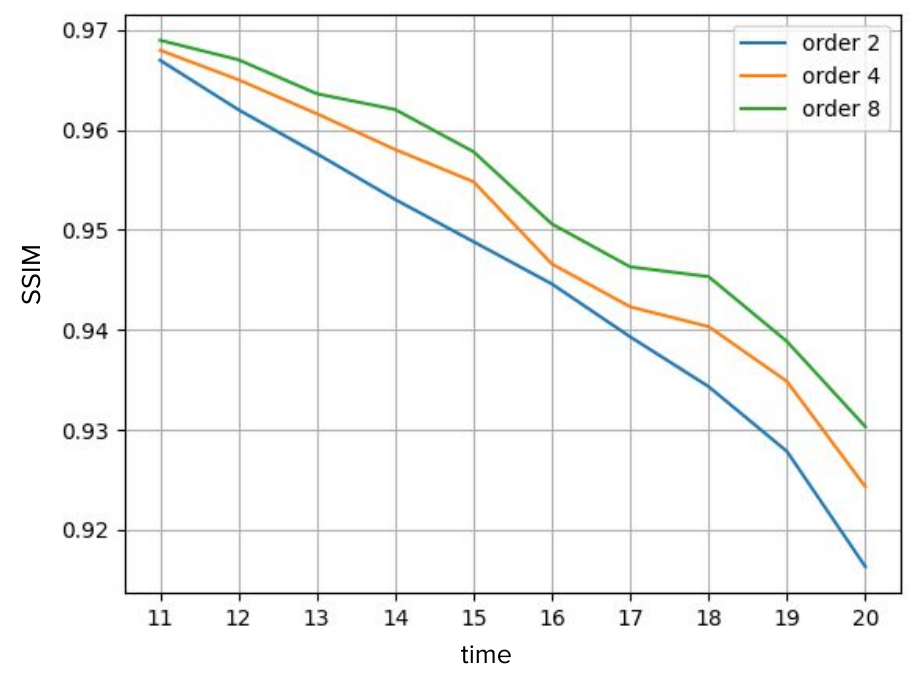} 
    \label{fig:compare_ty_orders}
}
\subfloat[][]{
\centering
    \includegraphics[trim={0 0 0 5pt}, clip, width=0.42\columnwidth]{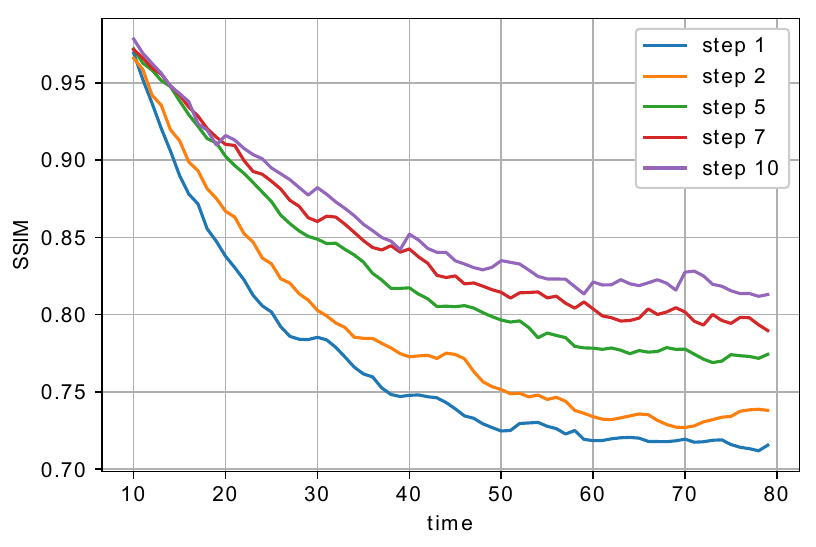}
    \label{fig:longterm}
}
\caption{(a) Comparing different orders of our temporal model using 10 frames as observation to predict the next 10 frames. (b) Comparison of future predictions for time horizons that are larger than the 10 frames used for training. The step size indicates after how many frames the Taylor approximation is performed. Step size $x$ means that our model forecasts $x$ frames, adds the forecast frames to the observations, and continues to predict the next $x$ frames. For both plots, we reduced the number of training epochs compared to the other experiments. 
}
\end{figure}

\textbf{Impact of $\gamma$.}
We approximate $\mathcal{F}_{\mathcal{H}_t}(t+\tau)$ \eqref{eq:taylor} by $\gamma$ terms where $\gamma$ defines the order of the Taylor series. 
We, therefore, evaluate the effect of using different orders of the Taylor series on Moving MNIST. As it can be seen in Fig.~\ref{fig:compare_ty_orders}, all three models have approximately the same prediction performance for $\tau{=}1$ (time step 11). However, as we increase $\tau$ the difference between the prediction performance of the 3 models increases. This is expected since having higher orders of the series will result in a better approximation of $\mathcal{F}_{\mathcal{H}_t}(t+\tau)$.

\textbf{Forecasting at Different Temporal Resolutions.}
To evaluate the capability of our model for forecasting at a higher temporal resolution than the observations, we train our model on Moving MNIST but we sample only every second frame, i.e., we use 5 out of 10 frames as observation and 5 out of 10 frames for prediction during training. For inference, we also sample 5 out of 10 frames as observation, but we aim to forecast all 10 frames. With our model, we can do this directly by generating the frames for $\tau=0.5, 1, 1.5, \ldots, 5$ and we compare the forecast frames to the 10 ground-truth frames. Note that the model is trained on the reduced temporal resolution (0.5x). Therefore, $\tau=5$ corresponds to the future frame 10 of the original sequence. The results are reported in Table~\ref{tab:temporal_res_expr}. For comparison, we generate the frames only for $\tau=1, 2, \ldots, 5$ as during training and linearly interpolate between the frames to get 10 frames. We also compare our approach to the state-of-the-art approach SimVP~\cite{Gao2022}. Since the sampling rate of SimVP cannot be changed after training, we use interpolation as well to get 10 frames. Our model achieves a much higher SSIM compared to approaches that require interpolation, which shows the benefit of a continuous representation that allows forecasting frames at a different temporal resolution than the observations.

\begin{table*}[t]
\centering
\resizebox{\columnwidth}{!}{
    \begin{tabular}{|c|c|c|c|c|c|c|}
         \hline
         & \multicolumn{4}{c|}{TaylorSwiftNet} & \multicolumn{2}{c|}{SimVP~\cite{Gao2022}}\\
         \cline{2-7}
         & Observed & Future & Observed & Future & Observed & Future \\
         Temporal Resolution & 0.5x & 1x  & 0.5x & 1x (Interp.) & 0.5x & 1x (Interp.) \\
         \hline
         SSIM $\uparrow$ & \multicolumn{2}{c|}{ \textbf{0.819}} & \multicolumn{2}{c|}{ 0.761}  &  \multicolumn{2}{c|}{0.779} \\
         \hline
    \end{tabular}
}    
\caption{Results when the model is trained on lower temporal resolution (0.5x). While the observed frames are also sub-sampled, the future frames are predicted at full temporal resolution (1x). In the case of `Interp.', the frames are predicted at a lower temporal resolution (0.5x), but upsampled by interpolation. Note that the sampling rate of the state-of-the-art method~\cite{Gao2022} cannot be changed and a higher frame-rate can only be achieved by interpolation.}
    \label{tab:temporal_res_expr}
\end{table*}

\textbf{Long-term Forecasting.}
In this experiment, we explore the long-term future forecasting capability of our TaylorSwiftNet. We use the same setup as in the previous experiments for the Moving MNIST dataset, but instead of predicting for the future temporal horizon of 10 frames we predict 70 frames, i.e., evaluating far beyond the prediction range seen during training. We, therefore, evaluate our model in a partially auto-regressive mode where we directly forecast the first 10 frames and feed them back to predict the next 10 frames. In other words, we do a Taylor approximation every 10 frames. For comparison, we also perform the Taylor approximation every 7, 5, 2, and each frame. The latter is a standard auto-regressive setting.   

As it can be seen in Fig.~\ref{fig:longterm}, performing the Taylor approximation every 10 frames performs best for forecasting longer sequences of future frames. In contrast, the accuracy of the standard auto-regressive setting where the frames are predicted frame-by-frame performs worst. The reason for such a fast drop in accuracy is due to the error propagation through the recursive steps. The models with the longer step size need fewer auto-regressive steps while the models with shorter step sizes need more auto-regressive steps. This shows that approaches that forecast frames frame-by-frame suffer from error propagation and are not suitable to forecast longer sequences.

\section{Conclusion}
In this work, we presented an approach that forecasts future frames by modeling the dynamics in the continuous time space without requiring any discretization. Since the motion can be very complex in a video, we use the Taylor series as the approximation method and train a network to infer the higher order terms of the Taylor series from the observed frames. We evaluated our approach on four datasets from different domains like forecasting human motion, hourly taxi flows, or sea surface temperature. For all datasets, our approach achieves state-of-the-art results. We also demonstrated that our approach is capable of forecasting frames at a higher temporal resolution than the observations. The approach, however, has some limitations. The resolution of the images is low and the predicted images become blurry for long-term predictions. The latter can be alleviated by using an additional adversarial loss to ensure that the images remain sharp.

\paragraph{Acknowledgement}
The work has been supported by the Deutsche Forschungsgemeinschaft (DFG, German Research Foundation) - SFB 1502/1-2022 - Project 450058266 and GA 1927/4-2 (FOR 2535 Anticipating Human Behavior), and the ERC Consolidator Grant FORHUE (101044724).

\appendix
\section*{Appendix}
In the following sections, we present additional ablation studies, more details of our method, and a comparison to analytical derived derivatives.

\section{Additional Ablation Studies}

\subsection{Recurrent DCBs}
As mentioned in the paper, the DC blocks can also be implemented as recurrent DCBs (RDCB). In contrast to DC blocks, RDC blocks share their weights. As shown in Table~\ref{tab:RDCBs}, DC blocks perform slightly better. We also found that DC blocks are more stable during training.
\begin{table}[h]
\begin{center}
        \begin{tabular}{|c|c|c|c|c|}
            \hline
             Method & Moving MNIST & Traffic BJ & SST & Human 3.6M  \\
            \hline 
            DCB & 0.965 & 0.992 & 0.978 & 0.910 \\
            \hline
            RDCB & 0.964 & 0.971 & 0.977 & 0.906\\
            \hline
        \end{tabular}    
\end{center}
\caption{SSIM for models with DCBs and RDCBs.}
\label{tab:RDCBs}
\end{table}

\subsection{Runtime Comparison}
We report the runtime of our model for Human 3.6M in Tab.~\ref{tab:runtime}. We used an NVIDIA Titan RTX GPU.

\begin{table}[h]
\begin{center}
        \begin{tabular}{|c|c|c|c|c|}
            \hline
              & run-time (ms) & Parameters (M) & GMACs & SSIM  \\
            \hline
              PhyDNet \cite{PhyDNet} & 30 & 11 & 76 & 0.901\\
            \hline
              ours & \textbf{21} & 11 & \textbf{61} & \textbf{0.910}\\
            \hline
        \end{tabular}
\end{center}
\caption{Computation cost and runtime.}
    \label{tab:runtime}
\end{table}

\begin{table}[h]
\begin{center}
        \begin{tabular}{|c|c|c|c|c|}
            \Xhline{2\arrayrulewidth}
             & $1^{st}$ & $2^{nd}$ & $3^{rd}$ & $4^{th}$  \\
            \hline 
            $d^i \sin/dt^i$ & $0.05077$ & $-0.99871$ & $-0.05077$ & $0.99871$ \\
            ours & $0.05083$ & $-0.99993$ & $-0.05017$ & $0.99012$\\
            \hline
            $d^i \cos/dt^i$ & $-0.9987$ & $-0.0507$ & $0.9987$ & $0.0507$ \\
            ours & $-0.9991$ & $-0.0506$ & $0.9900$ & $0.0504$\\
            \hline
            $d^i \exp/dt^i$ & $4.5722$ & $4.5722$ & $4.5722$ & $4.5722$ \\
            ours & $4.5723$ & $4.5726$ & $4.5721$ & $4.5730$\\
            \Xhline{2\arrayrulewidth}
        \end{tabular}    
\end{center}
\caption{Comparing the $1^{st}$, $2^{nd}$, $3^{rd}$, and $4^{th}$ order derivatives of three functions with the estimated derivatives.}
\label{tab:derivative_estimation}
\end{table}

\section{Comparison to Analytical Derivatives}
For video data, the function $\mathcal{F}_{\mathcal{H}_t}$ and thus the ground-truth terms of the Taylor series are unknown. In order to analyze how accurately our network can learn the terms of the Taylor series, we use three functions where we can analytically derive the derivatives. The results in Table \ref{tab:derivative_estimation} demonstrate that the DC blocks are able to learn derivatives. 

\section{Implementation Details}
\subsection{Encoder and Decoder}
\begin{figure*}[h!]
\begin{center}
\includegraphics[width=0.9\linewidth]{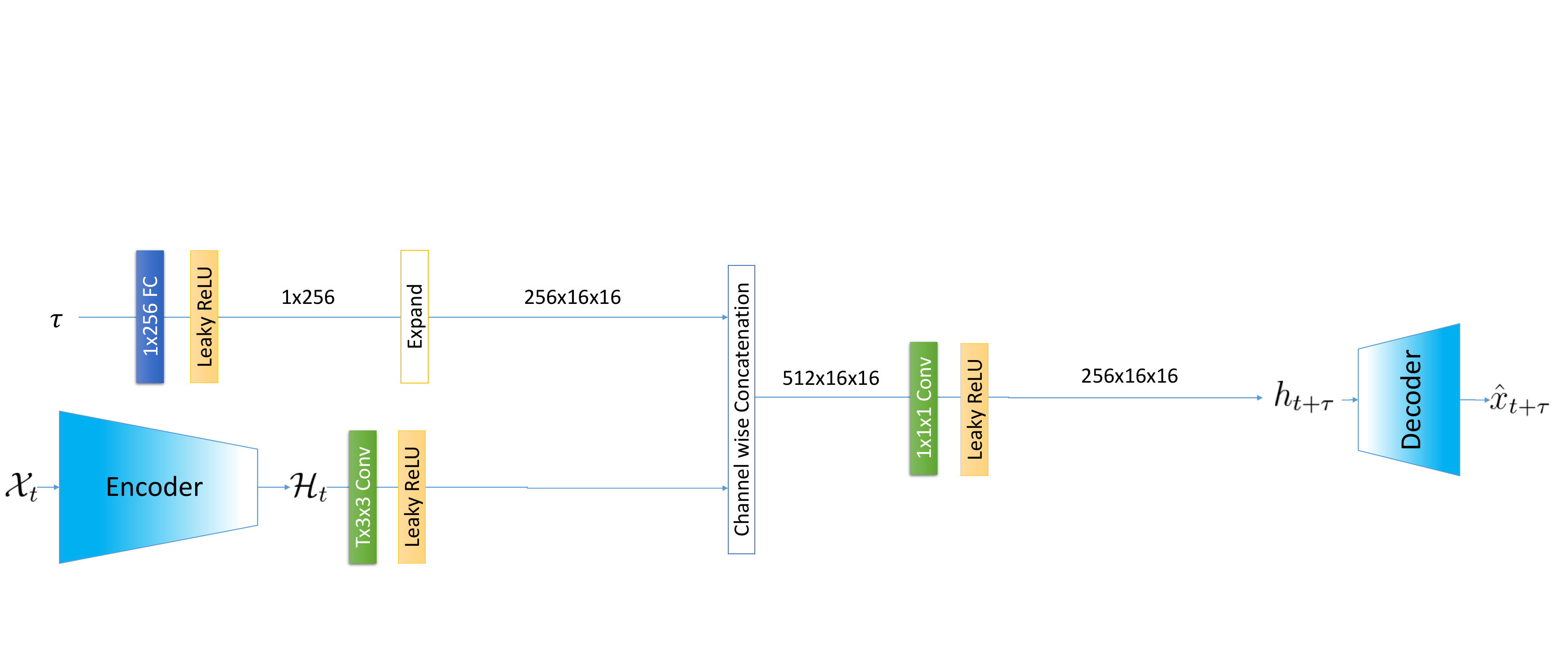}    
\end{center}
\caption{Architecture of the baseline `Expand'.}
\label{fig:Baseline2}
\end{figure*}

\begin{figure*}[h!]
\begin{center}
\includegraphics[width=0.9\linewidth]{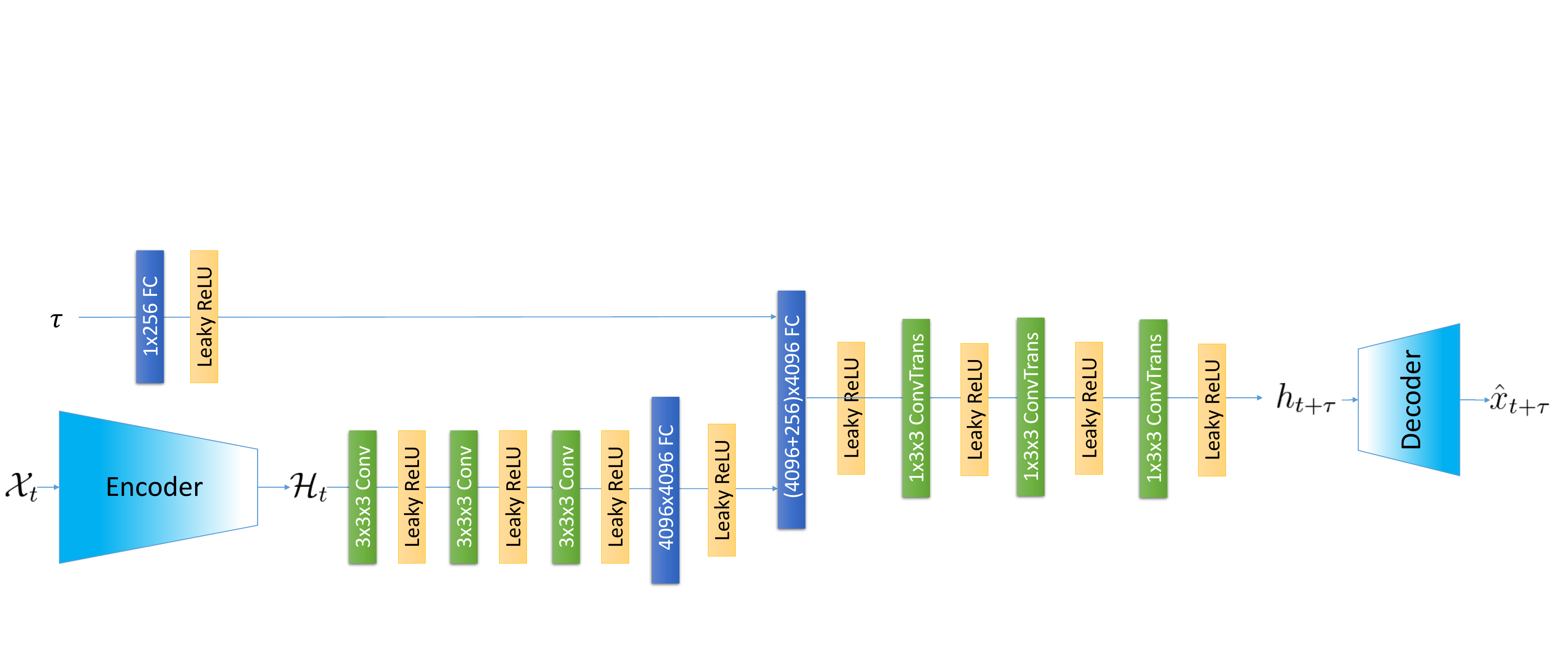}    
\end{center}
\caption{Architecture of the baseline `Flatten'.}
\label{fig:Baseline1}
\end{figure*}

We provide the details of the encoder and decoder for each dataset in Tables \ref{tab:moving_mnist_architecture}-\ref{tab:architecture_human}. The models share common blocks. We define each convolutional layer as: [input channel, output channel], [kernel height, kernel width, kernel depth], [stride over height, stride over width, stride over depth]. We also define each residual block as:
$$
\begin{aligned}
    \centering
    ResBlock = \begin{bmatrix}
    [C, C], [3, 3, 3], [1, 1, 1]
    \\
    [C, C], [3, 3, 3], [1, 1, 1]
    \end{bmatrix} 
\end{aligned}
$$
Furthermore, Figures \ref{fig:Baseline2} and \ref{fig:Baseline1} visualize the baselines `Point Estimate (Expand)' and `Point Estimate (Flatten)' from Table 2 of the paper.  

\subsection{Training}
We use Adam \cite{KingmaB14} with a learning rate of 0.0001 to optimize the model through 4K epochs. For the SST dataset, we train our model in 1K epochs. To control the learning rate, we use a scheduler to reduce the learning rate by a factor of 0.5 in case of a plateau over the SSIM metric on the training set. Following the previous state-of-the-art methods \cite{PhyDNet}, we use MSE as the loss function.

\section{Forecasting at Different Temporal Resolutions}
\begin{figure*}[h!]
\begin{center}
\includegraphics[width=1.0\linewidth]{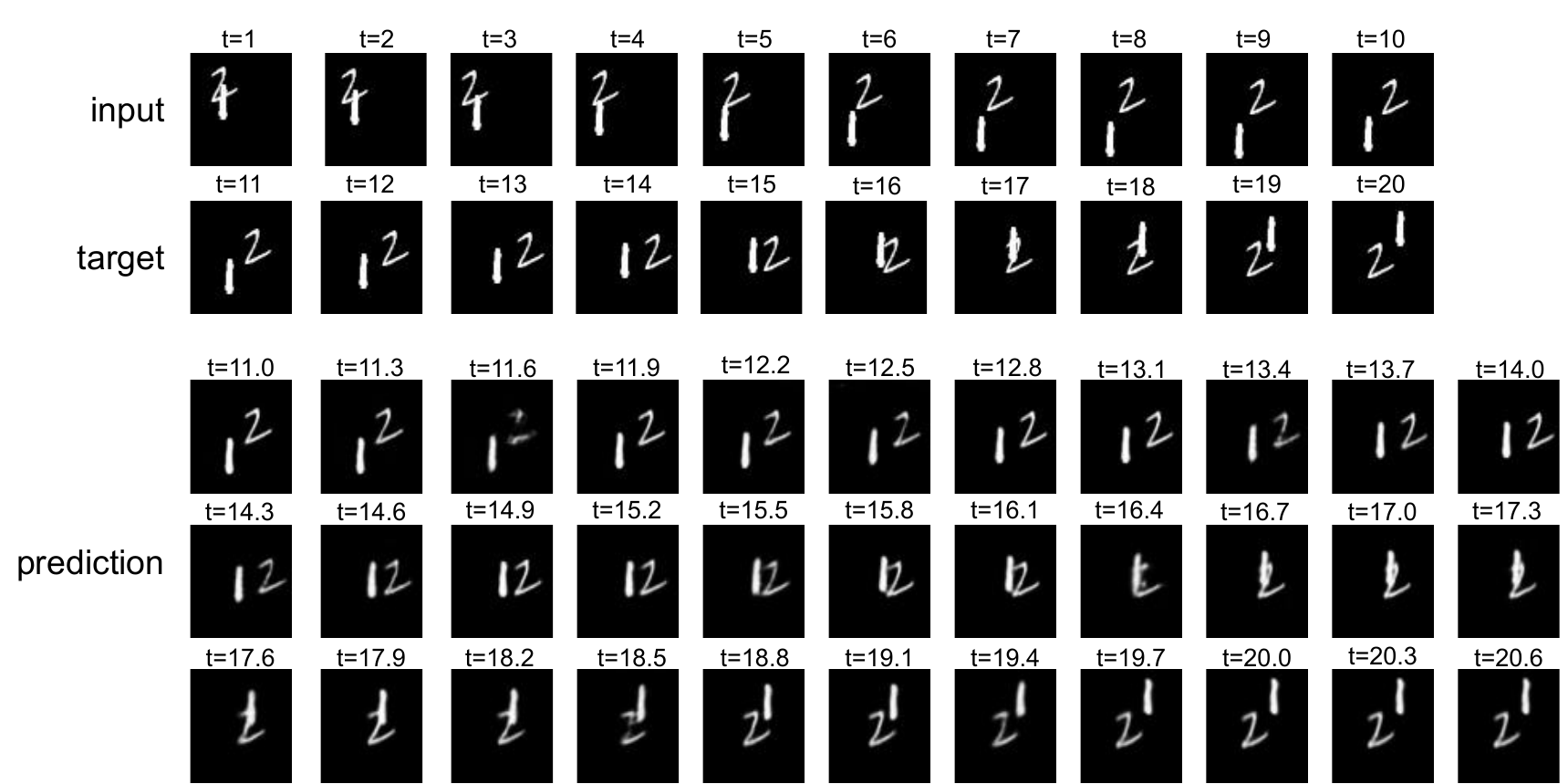}    
\end{center}
\caption{Predicting future frames at higher temporal resolution. Input is the 10 observed frames and the target are the future ground-truth frames. $\tau$ is increased by 0.3 instead of 1.}
\label{fig:continuous}
\end{figure*}

\begin{figure*}[h!]
\begin{center}
\includegraphics[width=1.0\linewidth]{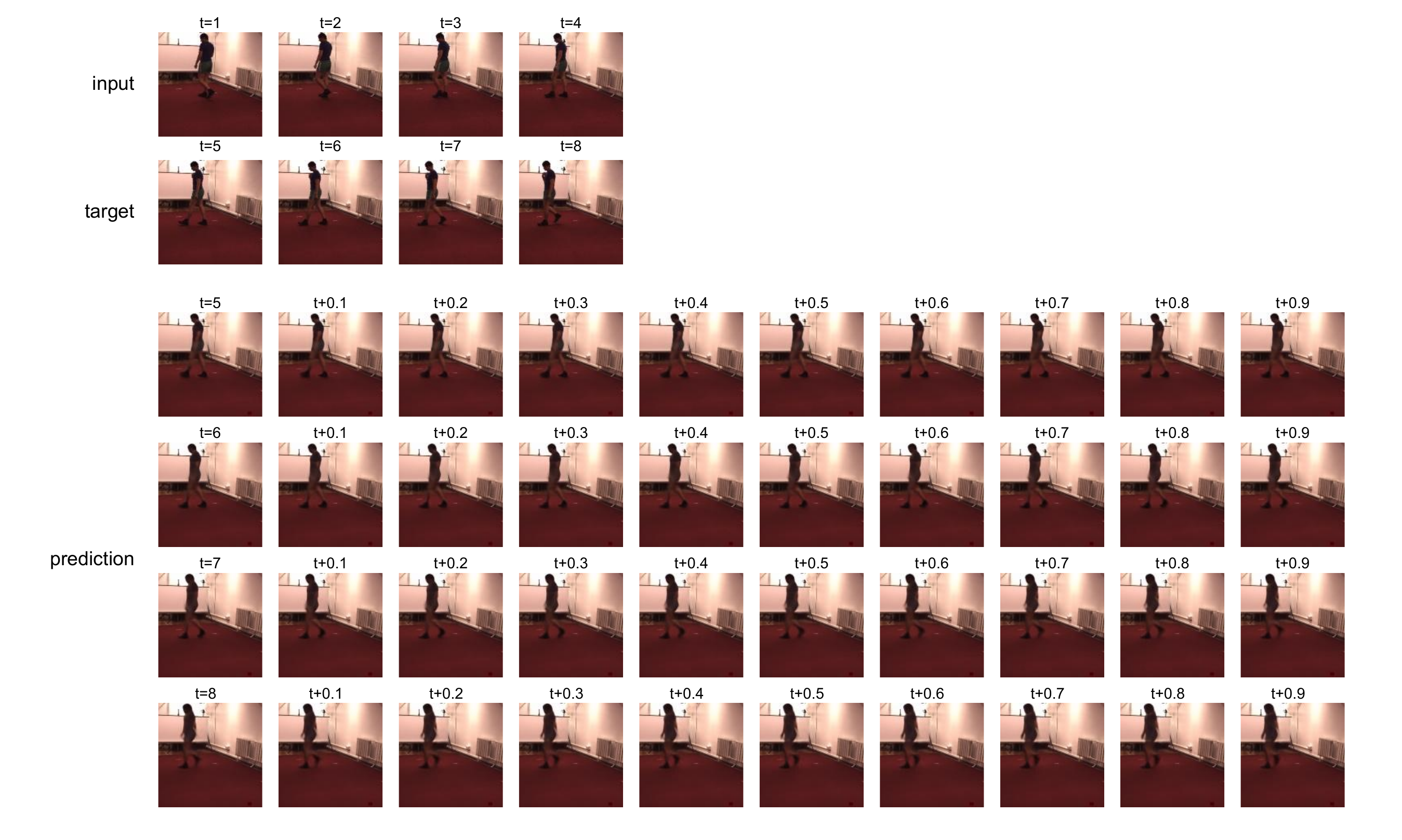}    
\end{center}
\caption{Predicting future frames at higher temporal resolution. Input is the 4 observed frames and the target are the future ground-truth frames. $\tau$ is increased by 0.1 instead of 1.}
\label{fig:human_interpolation}
\end{figure*}

Since our model forecasts frames using a continuous representation, we do not need to stick to the framerate of the observation. In Fig.~\ref{fig:continuous}, we show qualitative results on Moving MNIST for the future temporal steps $t{+}\tau \in \{11, 11.3, 11.6, ...,20.6\}$, i.e., we increase the framerate by $1/0.3$. Note that we do not re-train our model for this experiment. As it can be seen, our TaylorSwiftNet smoothly predicts intermediate frames. The sharp digits and their accurate location clearly demonstrate the continuous temporal modeling capability of our model.
We provide more qualitative results of the continuous temporal modeling capability of our method for Human 3.6M in Fig.~\ref{fig:human_interpolation}.

{\small
\bibliography{egbib}
}

\newpage 


\newpage 

\begin{table}[h!] 
\begin{center}
 \resizebox{0.5\columnwidth}{!}{%
        \begin{tabular}{c|c|c}
          stage & layer & output size \\
          \hline \hline 
          raw & - & $1\times10\times64\times64$ \\
          \hline
          \multirow{11}{*}{\rotatebox[origin=c]{270}{Encoder}} & [\ \ 1, \ \ 8], [1, 3, 3], [1, 1, 1] & $8\times10\times64\times64$ \\ \cline{2-3}
          & [\ \ 8, \ 16], [1, 3, 3], [1, 1, 1] & $16\times10\times64\times64$ \\ \cline{2-3}
          & [\ 16, \ 32], [1, 3, 3], [1, 2, 2] & \multirow{2}{*}{$32\times10\times32\times32$} \\ \cline{2-2}
          & [\ 32, \ 32], [1, 3, 3], [1, 1, 1] &  \\ \cline{2-3}
          & [\ 32, \ 64], [1, 3, 3], [1, 2, 2] & $64\times10\times16\times16$ \\ \cline{2-3}
          & [\ 64, 128], [1, 3, 3], [1, 1, 1] & \multirow{6}{*}{$128\times10\times16\times16$} \\ \cline{2-2}
          & [128, 128], [3, 3, 3], [1, 1, 1]&  \\ \cline{2-2}
          & ResBlock$_{2} \times 2$& \\ \cline{2-2}
          & ResBlock$_{3}\times 2$& \\ \cline{2-2}
          & ResBlock$_{4}\times 2$& \\ \cline{2-2}
          & ResBlock$_{5}\times 2$& \\ 
          \hline
          \multirow{6}{*}{\rotatebox[origin=c]{270}{Decoder}} & 
	      [128, 64], [1, 3, 3], [1, 1, 1] & $64\times10\times16\times16$ \\ \cline{2-3}
	      & [64, 32], [1, 3, 3], [1, 2, 2] & \multirow{2}{*}{$32\times10\times32\times32$} \\ \cline{2-2}
	      & [32, 32], [1, 3, 3], [1, 1, 1] &  \\ \cline{2-3}
	      & [32, 16], [1, 3, 3], [1, 2, 2] & $16\times10\times64\times64$ \\ \cline{2-3}
	      & [16, \ 8], [1, 3, 3], [1, 1, 1] & $8\times10\times64\times64$ \\ \cline{2-3}
	      & [\ 8, \ 1], [1, 3, 3], [1, 1, 1] & $1\times10\times64\times64$ \\
          \hline
        \end{tabular} 
        }
\end{center}
\caption{Model architecture with a modified 3DResNet encoder for Moving MNIST.}
\label{tab:moving_mnist_architecture}
\end{table}

\begin{table}[h!]
\begin{center}
 \resizebox{0.5\columnwidth}{!}{%
    \begin{tabular}{c|c|c}
          stage & layer & output size \\
          \hline \hline 
          raw & - & $2\times4\times64\times64$ \\
          \hline
          \multirow{8}{*}{\rotatebox[origin=c]{270}{Encoder}} & [\ \ 2, \ 32], [1, 3, 3], [1, 1, 1] & $32\times4\times32\times32$ \\ \cline{2-3}
          & [\ 32, \ 64], [1, 3, 3], [1, 2, 2] & $64\times4\times16\times16$ \\ \cline{2-3}
          & [\ 64, 128], [1, 3, 3], [1, 1, 1] & \multirow{6}{*}{$128\times4\times16\times16$} \\ \cline{2-2}
          & [128, 128], [3, 3, 3], [1, 1, 1] &  \\ \cline{2-2}
          & ResBlock$_{2} \times 2$& \\ \cline{2-2}
          & ResBlock$_{3}\times 2$& \\ \cline{2-2}
          & ResBlock$_{4}\times 2$& \\ \cline{2-2}
          & ResBlock$_{5}\times 2$& \\ 
          \hline
	      \multirow{3}{*}{\rotatebox[origin=c]{270}{Decoder}} & 
	      [128, 64], [1, 3, 3], [1, 1, 1] & $64\times4\times16\times16$ \\ \cline{2-3}
	      & [\ 64, 32], [1, 3, 3], [1, 2, 2] & $32\times4\times32\times32$ \\ \cline{2-3}
	      & [\ 32, \ \ 2], [1, 3, 3], [1, 1, 1] & $2\times4\times32\times32$ \\
          \hline
        \end{tabular}
        }
\end{center}
\caption{Model architecture with a modified 3DResNet encoder for Traffic BJ.}
\label{table:traffic_architecture}
\end{table}

\newpage

\begin{table}[h!]
\begin{center}
 \resizebox{0.5\columnwidth}{!}{%
        \begin{tabular}{c|c|c}
          stage & layer & output size \\
          \hline \hline 
          raw & - & $1\times4\times64\times64$ \\
          \hline
          \multirow{10}{*}{\rotatebox[origin=c]{270}{Encoder}} & [\ \ 1, \ 16], [1, 3, 3], [1, 1, 1] & $16\times4\times64\times64$ \\ \cline{2-3}
          & [\ 16, \ 32], [1, 3, 3], [1, 2, 2] & \multirow{2}{*}{$32\times4\times32\times32$} \\ \cline{2-2}
          & [\ 32, \ 32], [1, 3, 3], [1, 1, 1] & \\ \cline{2-3}
          & [\ 32, \ 64], [1, 3, 3], [1, 2, 2] & $64\times4\times16\times16$ \\ \cline{2-3}
          & [\ 64, 128], [1, 3, 3], [1, 1, 1] & \multirow{6}{*}{$128\times4\times16\times16$} \\ \cline{2-2}
          & [128, 128], [3, 3, 3], [1, 1, 1] &  \\ \cline{2-2}
          & ResBlock$_{2} \times 2$& \\ \cline{2-2}
          & ResBlock$_{3}\times 2$& \\ \cline{2-2}
          & ResBlock$_{4}\times 2$& \\ \cline{2-2}
          & ResBlock$_{5}\times 2$& \\ 
          \hline
	      \multirow{5}{*}{\rotatebox[origin=c]{270}{Decoder}} & 
	      [128, 64], [1, 3, 3], [1, 1, 1] & $64\times4\times16\times16$ \\ \cline{2-3}
	      & [64, 32], [1, 3, 3], [1, 2, 2] & \multirow{2}{*}{$32\times4\times32\times32$} \\ \cline{2-2}
	      & [32, 32], [1, 3, 3], [1, 1, 1] & \\ \cline{2-3}
	      & [32, 16], [1, 3, 3], [1, 2, 2] & $16\times4\times64\times64$ \\ \cline{2-3}
	      & [16, 1], [1, 3, 3], [1, 1, 1] & $1\times4\times64\times64$ \\
          \hline
          \end{tabular}
}
\end{center}
\caption{Model architecture with a modified 3DResNet encoder for SST.}
\label{tab:architecture_sst}
\end{table}
 
 \begin{table}[h!] 
 \begin{center}
 \resizebox{0.5\columnwidth}{!}{%
        \begin{tabular}{c|c|c}
          stage & layer & output size \\
          \hline \hline 
          raw & - & $3\times4\times64\times64$ \\
          \hline         
          \multirow{11}{*}{\rotatebox[origin=c]{270}{Encoder}} & [3, 16], [1, 3, 3], [1, 1, 1] & $16\times4\times64\times64$ \\ \cline{2-3}
          & [16, 32], [1, 3, 3], [1, 1, 1] & $32\times4\times64\times64$ \\ \cline{2-3}
          & [32, 64], [1, 3, 3], [1, 2, 2] & \multirow{2}{*}{$64\times4\times32\times32$} \\ \cline{2-2}
          & [64, 64], [1, 3, 3], [1, 1, 1] & \\ \cline{2-3}
          & [64, 128], [1, 3, 3], [1, 2, 2] & $128\times4\times16\times16$ \\ \cline{2-3}
          & [128, 256], [1, 3, 3], [1, 1, 1] & \multirow{6}{*}{$256\times4\times16\times16$} \\ \cline{2-2}
          & [256, 256], [3, 3, 3], [1, 1, 1] &  \\ \cline{2-2}
          & ResBlock$_{2} \times 2$& \\ \cline{2-2}
          & ResBlock$_{3}\times 2$& \\ \cline{2-2}
          & ResBlock$_{4}\times 2$& \\ \cline{2-2}
          & ResBlock$_{5}\times 2$& \\ 
          \hline
	      \multirow{6}{*}{\rotatebox[origin=c]{270}{Decoder}} & 
	      [256, 128], [1, 3, 3], [1, 1, 1] & $128\times4\times16\times16$ \\ \cline{2-3}
	      & [128, 64], [1, 3, 3], [1, 2, 2] & \multirow{2}{*}{$64\times4\times32\times32$} \\ \cline{2-2}
	      & [64, 64], [1, 3, 3], [1, 1, 1] & \\ \cline{2-3}
	      & [64, 32], [1, 3, 3], [1, 2, 2] & $32\times4\times64\times64$ \\ \cline{2-3}
	      & [32, 16], [1, 3, 3], [1, 1, 1] & $16\times4\times64\times64$ \\ \cline{2-3}
	      & [16, 3], [1, 3, 3], [1, 1, 1] & $3\times4\times64\times64$ \\
          \hline
        \end{tabular} 
        }
 \end{center}
\caption{Model architecture with a modified 3DResNet encoder for Human 3.6M.}
\label{tab:architecture_human}
\end{table}

\end{document}